\documentclass[runningheads]{llncs}

\usepackage{eccv}

\usepackage{eccvabbrv}

\usepackage{graphicx}
\usepackage{booktabs}
\usepackage{tcolorbox}

\usepackage[accsupp]{axessibility}  

\usepackage{hyperref}

\usepackage{orcidlink}

\usepackage{tabularx}
\usepackage{amsmath}
\usepackage{makecell}
\usepackage{array} 
\usepackage{multirow}
\usepackage{lipsum}
\usepackage{booktabs}
\usepackage{color}
\usepackage{colortbl}
\usepackage{subcaption}

\definecolor{turquoise}{cmyk}{0.65,0,0.1,0.3}
\definecolor{purple}{rgb}{0.65,0,0.65}
\definecolor{dark_green}{rgb}{0, 0.5, 0}
\definecolor{orange}{rgb}{0.8, 0.6, 0.2}
\definecolor{red}{rgb}{0.9, 0.1, 0.1}
\definecolor{darkred}{rgb}{0.6, 0.1, 0.05}
\definecolor{blueish}{rgb}{0.0, 0.3, .6}
\definecolor{light_gray}{gray}{0.95}
\definecolor{pink}{rgb}{1, 0, 1}
\definecolor{greyblue}{rgb}{0.25, 0.25, 1}

\newcommand{\ul}[1]{\underline{#1}}

\newcommand{\paragrapht}[1]{\noindent\textbf{#1}}  

\usepackage{pifont}
\usepackage{amssymb}

\usepackage{xspace}
\usepackage{array}
\usepackage{layouts}
\usepackage{algorithm}
\usepackage{bbding}
\usepackage{listings}
\usepackage{pifont}
\usepackage{wasysym}
\usepackage{float}
\usepackage{footnote}
\makesavenoteenv{tabular}
\makesavenoteenv{table}
\makesavenoteenv{figure}
\usepackage{pythonhighlight}
\usepackage{bbm}
\usepackage{wrapfig}

\newcommand{\ours}{\textsc{Splash}\xspace}
\newcommand{\oursfull}{{Mask-guided tactile alignment learning}\xspace}
\newcommand{\llm}{\texttt{LLM}_{\theta}\xspace}
\newcommand{\rgbenc}{\mathcal{F}_{\phi}\xspace}
\newcommand{\tacenc}{\mathcal{F}_{\psi}\xspace}

\newcommand{\weight}{\mathbf{W}}
\newcommand{\score}{\mathbf{S}}
\newcommand{\act}{\mathbf{x}}
\newcommand{\onemask}{\mathbbm{1}}

\begin{document}

\title{Wake up for Touch! Mask-isolated Tactile Alignment Learning in MLLMs} 

\titlerunning{Wake up for Touch! Mask-isolated Tactile Alignment Learning in MLLMs}

\author{Yoonhyung Park$^*$ \and
Minji Kim$^*$ \and
Sungwon Moon \and
Jiyoung Lee$^\dagger$}

\authorrunning{Y.~Park and M.~Kim et al.}

\institute{Division of Artificial Intelligence \& Software, Ewha Womans University \\ \vspace{.5em}
\email{\{pyoon0820, xxinzzi03\}@ewhain.net,  \{sungwon268,lee.jiyoung\}@ewha.ac.kr}\\ \vspace{.5em}
\url{http://mmai.ewha.ac.kr/splash/}}

\maketitle
\def\thefootnote{*}\footnotetext{These authors contributed equally to this work.}\def\thefootnote{\arabic{footnote}}
\def\thefootnote{$\dagger$}\footnotetext{Corresponding author.}\def\thefootnote{\arabic{footnote}}

\begin{abstract}
    Touch supplies the physical grounding needed to perceive intrinsic material properties, such as friction and compliance, that vision alone often cannot resolve. 
    Recent efforts for equipping multimodal LLMs with this tactile sense, however, expose a zero-sum trade-off: the limited parameter budget of compact models forces a choice between acquiring the new sensory modality and preserving the established vision-language reasoning.
    We present \textbf{\ours}, a mask-isolated tactile alignment learning framework for MLLMs. 
    \ours quantifies the significance of each pretrained parameter, and partitions the parameter space into a dormant and critical subspace.
    While the frozen critical subspace acts as a stable anchor to safeguard general visual knowledge, \ours updates the isolated dormant subspace to internalize tactile alignment towards LLMs.
    This selective, non-destructive expansion effectively prevents catastrophic forgetting and ensures non-destructive modality expansion.
    Extensive experiments show that \ours effectively achieves tactile reasoning without additional inference overhead in the LLM part, demonstrating state-of-the-art performance on visuo-tactile benchmarks, including SSVTP, TVL, and TacQuad, while preserving its original general-purpose capabilities.
\end{abstract}
\section{Introduction}\label{sec:intro}
Humans interact with the physical world not merely by looking at it, but by touching or pressing it to judge whether a surface is slippery, compliant, or rigid.
This innate ability to perceive intrinsic material properties such as friction and compliance is essential feedback for precise physical interaction.
For embodied robots to achieve similar dexterity, perceiving fine-grained contact dynamics is crucial for manipulating objects.
However, vision-based systems often struggle to infer such properties, which are frequently occluded or visually indistinguishable during active manipulation~\cite{calandra2018more,yuan2017connecting}.

To close this gap, recent approaches~\cite{yang2022touch, cheng2025tlvlink, feng2025anytouch, Lei_2024_CVPR} have explored learning cross-modal associations between visual appearance and tactile feedback. 
These approaches typically learn a shared visuo-tactile-language (VTL) representation space.  
While effective for discriminative tasks such as retrieval and classification, they lack the capacity for the higher-level semantic reasoning and instruction following that complex embodied decision-making demands. 
Moreover, their learned representations generalize poorly to novel objects with visually ambiguous textures.
These observations indicate the need for models capable of reasoning over multimodal sensory inputs.

Multimodal large language models (MLLMs)~\cite{llava,bai2023qwenvl,chen2024internvl} are a natural fit.
By drawing on the world knowledge encoded in their LLM backbones, MLLMs support open-vocabulary reasoning over sensory inputs, allowing robots not only to perceive but also to reason about the physical consequences of their actions.
Most existing MLLMs, however, rely on computationally intensive backbones (\eg, LLaMA-7B~\cite{llama}), which limit their practicality for resource-constrained edge robots~\cite{williams2025lite}. 
This motivates compact MLLMs capable of tactile reasoning under real-world deployment constraints~\cite{chu2024mobilevlm}.

\begin{figure}[t]
    \centering
    \includegraphics[width=\linewidth]{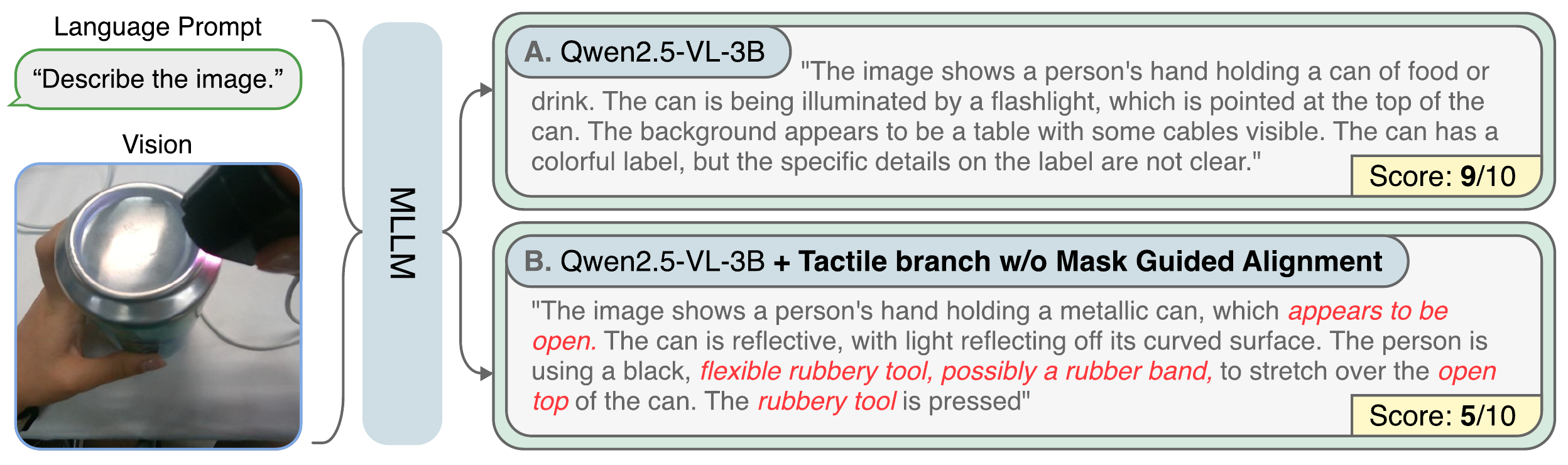}
    \caption{Example of the catastrophic forgetting problem in tactile alignment for MLLMs (\eg, TVL~\cite{tvl} w/ Qwen2.5-VL-3B). A small amount of tactile training set often forgets the visual sense in the base MLLM. More failure cases in Appendix.}
    \label{fig:forgetting}
\end{figure}

The challenge is most acute for small MLLMs (sMLLMs, \eg, $\leq$ 3B-scale), which are the ones actually desirable for edge deployment. Their limited capacity means that incorporating a new touch modality interferes with existing vision-language representations. 
As \cref{fig:forgetting} illustrates, optimizing such a model for tactile signals distorts its pretrained visual features, causing catastrophic forgetting and degraded vision-language (VL) reasoning performance. 
The result is a fundamental trade-off between expanding sensory capabilities and preserving the pretrained reasoning abilities of sMLLMs.

In this paper, we propose \ours, a novel framework that aligns tactile signals into sMLLMs while mitigating catastrophic forgetting~\cite{mccloskey1989catastrophic, kirkpatrick2017overcoming, zhai2023forgetting}. 
The key idea is to confine all tactile adaptation to a dormant (\ie, less important) parameter subspace while keeping the critical parameters frozen. 
We estimate parameter importance with a VL-relative metric computed from weight and activation statistics, then restrict gradient updates to the dormant subspace. 
Since frozen critical parameters are stable vision-language anchors, our selective optimization approach preserves pretrained reasoning capabilities while enabling the model to incorporate tactile information. 
By isolating and reactivating internal weights, \ours maintains the original MLLM scale without introducing external modules like adapters, aside from a negligible tactile front-end.
Moreover, \ours supports a unified, single-stage training, without multi-stage alignment and fine-tuning procedures \cite{tvl, unitouch}.
Extensive experiments on visuo-tactile-language (VTL) benchmarks, including SSVTP~\cite{ssvtp}, TVL~\cite{tvl}, and TacQuad~\cite{feng2025anytouch}, validate the effectiveness of \ours. 
Notably, \ours-3B achieves state-of-the-art (SOTA) performance, outperforming previous 7B-scale methods~\cite{tvl, unitouch}.
In addition, we present the first comprehensive performance analysis of preservation in VTL alignment across complex general-purpose VL benchmarks.
The results demonstrate that tactile alignment with \ours does not compromise the broad-spectrum intelligence of MLLM backbones.

We summarize our contributions as follows:
\begin{enumerate}
    \item We introduce \ours, a parameter-isolated tactile adaptation framework that integrates tactile sensing into sMLLMs with zero additional inference overhead, mitigating the vision-related catastrophic forgetting.
    \item By leveraging frozen critical parameters as stable structural anchors, \ours enables single-stage unified training that optimizes tactile front-end and the dormant LLM subspace concurrently.
    \item Across small-scale  MLLMs including Qwen2.5-VL-3B and InternVL-1B, \ours achieves SOTA performance on three VTL benchmarks while safeguarding the models' pretrained general-purpose capabilities.
\end{enumerate}

\section{Related Work}
\label{sec:related}

\subsection{Visuo-Tactile-Language (VTL) Alignment}
The integration of vision and tactile sensing has long been central to robotic perception, chiefly as a way to resolve the ambiguities of any single modality.
Early work established the value of synergistic visuo-tactile integration across a spectrum of robotic tasks such as force estimation~\cite{chelly2025tactile,yuan2017gelsight}, cross-modal synthesis~\cite{yang2022touch,yuan2017connecting}, and robotic manipulation~\cite{calandra2018more, heng2025vitacformer}.
With the rise of MLLMs~\cite{llava,bai2023qwenvl,chen2024internvl}, the attention has shifted toward VTL alignment, where the tactile modality is aligned with a vision-language latent space, typically via contrastive learning objectives~\cite{yang2022touch, cheng2025tlvlink, unitouch}. 
Subsequent efforts push scalability further by constructing large-scale VTL datasets~\cite{cheng2025tlvlink, tvl} and by learning cross-sensor visuo-tactile representations~\cite{unitouch, feng2025anytouch, feng2026anytouch2generaloptical} that generalize across heterogeneous tactile sensors. 
Despite these advances, how to integrate tactile sensing into compact multimodal models remains largely underexplored. 
In this work, we address this gap directly, studying how the tactile modality can be folded into small MLLMs without sacrificing their pretrained abilities.

\subsection{Pruning-based Multi-Task Learning}
Our work shares the underlying principle of pruning-based multi-task learning~\cite{packnet, lotteryticket}, where disjoint parameter subsets are allocated to different tasks to mitigate cross-task interference.
Recent findings in LLMs~\cite{khaki2025sparselora, forgetting_aware_2025, frantar2023sparsegpt, ma2023llm} reinforce this view: updating only a small fraction of parameters can achieve competitive performance compared to full finetuning while effectively curbing catastrophic forgetting.
These results imply that model capacity can be selectively exploited without disturbing the entire parameter space. 
However, these works predominantly focus on intra-modality adaptation, such as continual learning~\cite{mccloskey1989catastrophic,kirkpatrick2017overcoming,li2017learning}, primarily aiming to pack multiple tasks into one model while minimizing interference among them.
We instead extend this principle from task-level to modality-level parameter isolation, addressing the severe representation gap encountered in new tactile sensory alignment. 

\section{\ours}
\label{sec:method}

\subsection{Motivation and Problem Formulation}
MLLMs~\cite{llama,li2023blip,zhu2023minigpt,chen2024internvl} typically consist of a vision encoder, a modality adapter, and an LLM that reasons over the fused multimodal input.
Of these, the LLM's scale largely governs reasoning capacity and generalization.
Enlarging the LLM improves performance on complex tasks and zero-shot generalization, but at a steep computational cost.
This escalation in model size presents a critical bottleneck for robotic applications, where real-time inference and limited on-board compute are required~\cite{zitkovich2023rt,driess2023palm,chu2024mobilevlm}.
Worse, naively integrating a new sensory modality risks catastrophic forgetting: tactile instruction tuning that updates the LLM parameters can overwrite the weight configurations encoding pretrained visual knowledge, even when only a small amount of tactile data is used.

To address these challenges, we propose \oursfull, \ours.
Given a natural (\ie, RGB) image $X_R$, a tactile contact image $X_C$, and a text prompt $X_T$, the goal is to generate a tactile description $Y$ in natural language form:
\begin{align}
    Y = \texttt{LLM}_{\theta}(\mathcal{F}_\phi(X_R), \mathcal{F}_\psi(X_C), X_T),
\end{align}
where $\theta$ are the language-model parameters. 
We denote the visual front-end, consisting of a vision encoder and modality adapter, as $\mathcal{F}$, with the natural-image and tactile front-ends parameterized by $\phi$ and $\psi$, respectively.
We initialize $\llm$ and $\rgbenc$ using weights from a pretrained sMLLM~\cite{Qwen2.5-VL} to inherit its general-purpose reasoning and visual understanding.
For the tactile front-end $\tacenc$, we employ an ImageNet-pretrained~\cite{deng2009imagenet} vision transformer (ViT)~\cite{dosovitskiy2021an} to extract geometric features from contact images.
During training, we freeze the visual parameters $\phi$ and update the tactile parameters $\psi$ together with a subset of the LLM parameters $\theta$, determined by a sparsity ratio $\tau$.
Next, we describe the key components of our proposed framework in detail. 

\subsection{Locating the Dormant Subspace}\label{sec:dormant}

\begin{figure}[t]
    \centering
    \includegraphics[width=\linewidth]{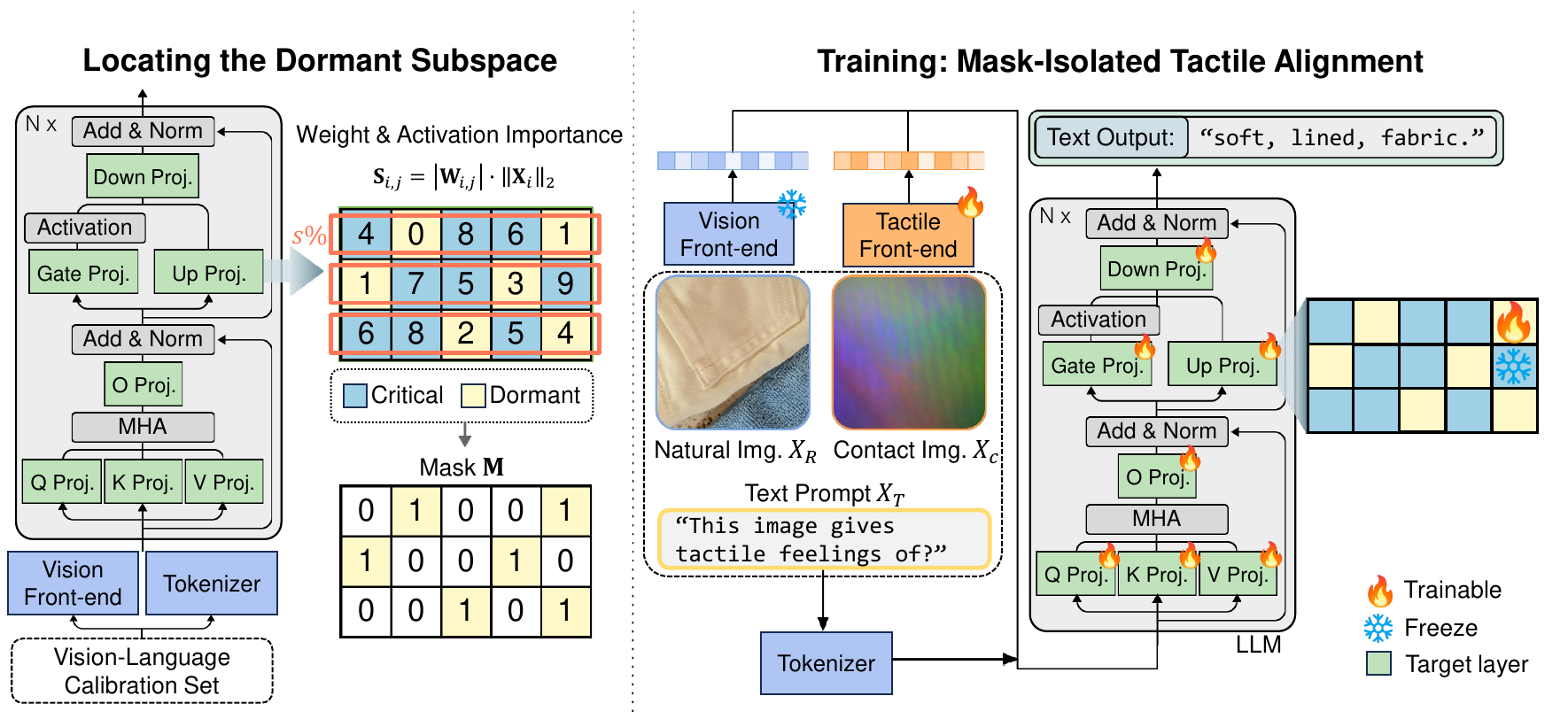}
    \caption{The overall framework of \ours. To mitigate the catastrophic forgetting in the visual aspect, we train a dormant subspace in LLM for tactile alignment in sMLLMs. We note that the tactile front-end is also updated in a unified training stage.}
    \label{fig:pipeline}
\end{figure}

Given a pretrained MLLM (\eg, QwenVL-2.5~\cite{Qwen2.5-VL}, InternVL~\cite{chen2024internvl}), \ours first identifies a dormant parameter subspace that contributes minimally to pretrained VL tasks. 
This builds on the \textit{natural sparsity} hypothesis~\cite{lotteryticket}, which holds that only a subset of parameters is critical for maintaining model function.
We exploit such redundancy by repurposing these underutilized weights, referred to as the \textbf{dormant subspace}, as trainable capacity for the tactile modality.

To identify dormant parameters, we use a visual-relative importance metric inspired by Wanda~\cite{sun2024wanda}.
With a small calibration set drawn from VL data (\eg, 128 samples from CC12M~\cite{changpinyo2021conceptual}), we score the importance of each weight $\weight_{i,j}$ connecting the $i$-th input feature to the $j$-th output neuron in the LLM's linear layers. 
The resulting scores reflect each parameter's contribution to pretrained VL behavior.
Formally, the importance score $\score_{i,j}$ is defined as:
\begin{align}
    \score_{i,j} = |\weight_{i,j}| \cdot \| \act_i \|_2,
\end{align}
where $|\cdot|$ denotes the weight magnitude and $\| \act_i \|_2$ denotes the $\ell_2$-norm of the corresponding input activation.
Given a sparsity ratio $s$\%, we set a threshold $\tau_j(s)$ such that $s$\% of the scores in $\{\score_{i,j}\}_{i=1}^{d_{\text{in}}}$ fall below it, and form the binary mask within each layer: 
\begin{equation}
\mathbf{M}_{i,j} = 
\onemask \Bigl( \score_{i,j} < \tau_j(s) \Bigr),
\end{equation}
where $\onemask(\cdot)$ denotes the indicator function.
For example, when $s=0\%$, all parameters are treated as critical and remain frozen, whereas $s=100\%$ corresponds to full-parameter tuning of the LLM.
We additionally set $\mathbf{M}_{i,j}=0$ for the first and last transformer blocks, preserving input grounding and high-level semantic representations.
It thereby ensures training stability and prevents representation collapse during tactile adaptation~\cite{sun2024wanda,alayrac2022flamingo}.
Consequently, the resulting mask cleanly partitions the LLM into frozen critical weights and trainable dormant weights, enabling tactile alignment while protecting the pretrained VL capabilities of the model.

\subsection{Mask-isolated Tactile Alignment}
Existing VTL alignment methods~\cite{tvl, feng2025anytouch} typically follow a two-stage training scheme. 
First, the tactile encoder is aligned with either the vision-language~\cite{tvl} or the vision representation space~\cite{feng2025anytouch}. 
Second, the LLM is adapted using parameter-efficient finetuning (PEFT) (\eg, LoRA~\cite{lora}, $\mathrm{(IA)}^3$~\cite{liu2022few}) for multimodal reasoning.
\ours instead enables a unified, single-stage training process by leveraging the parameter partition from \cref{sec:dormant}.
Specifically, we optimize the tactile front-end $\tacenc$ and the dormant LLM parameters while holding the critical parameters frozen.
The training objective $\mathcal{L}$ maximizes the likelihood of the autoregressively generated tactile description as follows:
\begin{equation}
\mathcal{L} = - \sum_{n=1}^N \log P(y_n | Y_{<n}, X_R, X_C, X_T),
\end{equation}
where $N$ is the number of tokens in the output sequence $Y$.
We apply the binary mask $\mathbf{M}$ to the LLM backbone parameters, while the tactile parameters $\psi$ remain fully trainable. 
Let $\eta$ denote the learning rate. 
The parameter updates are defined as:
\begin{align}
\theta^* \, \leftarrow \, & 
\theta
- \eta
\left(
\mathbf{M} \odot \nabla_{\theta} \mathcal{L}
\right), \label{eq:backbone_update}\\
\psi \, \leftarrow \, &
\psi
- \eta \nabla_{\psi} \mathcal{L}, \label{eq:tactile_update}
\end{align}
where $\odot$ denotes Hadamard multiplication.
\cref{eq:backbone_update} gives the critical parameters (\ie, $\mathbf{M}_{i,j}=0$) zero gradient and dormant parameters (\ie, $\mathbf{M}_{i,j}=1$) for gradient update, while \cref{eq:tactile_update} allows the tactile branch to train fully.

Given that we preserve the critical vision-language weights, these frozen parameters work as stable anchors that maintain the integrity of the multimodal manifold during adaptation.
Crucially, unlike additive PEFT methods such as LoRA ($\theta + \Delta \theta$), which can perturb the entire parameter manifold, \ours employs strict structural isolation by confining updates exclusively to the dormant subspace. 
This sparse replacement mechanism ensures that the pretrained VL reasoning pathways remain fundamentally untouched, effectively preventing catastrophic forgetting without any additional inference latency.
Therefore, this design eliminates the need for a separate alignment stage and simplifies the training pipeline, significantly reducing the overall training cost, and cohesively integrating the tactile into the reasoning process of MLLMs.
\section{Experiments}
\label{sec:experiments}

\subsection{Datasets}

\paragrapht{Visuo-Tactile-Language (VTL).}
For VTL alignment learning, we train \ours on DIGIT-sensor-based datasets, SSVTP~\cite{ssvtp} and TVL~\cite{tvl}.
SSVTP contains 4.5K aligned visuo-tactile pairs, whereas TVL contains 44K pairs annotated with natural language descriptions of tactile properties. 
We use the official train/test split and further divide the training set into train/validation at a 9:1 ratio.
To assess robustness under distribution shift,
we further evaluate on the DIGIT-sensor subset of TacQuad~\cite{feng2025anytouch} (72K tactile samples), collected via handheld interaction to capture diverse real-world contact dynamics of humans.
This dataset includes a detailed linguistic description, initially generated by GPT-4o~\cite{hurst2024gpt} and manually revised.
By leveraging this richly annotated dataset, our evaluation enables a more natural language-based assessment beyond the keyword-based evaluation.

\paragrapht{General-purpose Vision-Language (VL).} 
To assess retention of pretrained VL capabilities after tactile adaptation, we evaluate models on standard vision-language benchmarks, including MMMU~\cite{mmmu} for expert-level multi-disciplinary reasoning, MathVista~\cite{mathvista} for complex mathematical logic within visual context, MME~\cite{mme} for broad perception and cognitive stability, and MMBench~\cite{mmbench} for assessment on bilingual context.
These benchmarks evaluate diverse multimodal reasoning and perception capabilities, allowing us to assess whether tactile adaptation preserves the pretrained VL capabilities of the model.

\subsection{Evaluation Protocol}
Following the evaluation protocol of TVL~\cite{tvl}, we evaluate tactile semantic generation on SSVTP, TVL and TacQuad using an LLM-based evaluator (\ie, LLM-as-a-judge).  Since TacQuad consists of trajectory-level tactile and visual image sequences paired with one free-form textual description, we use the middle frame of each contact trajectory as the representative observation for evaluation.
Specifically, we adopt GPT-4o~\cite{hurst2024gpt} as a text-only judge to score the semantic similarity between the generated tactile descriptions and ground-truth tactile descriptions. 
The score ranges from 1 to 10, where higher scores indicate better semantic alignment. 
The score is computed as the average GPT-4o score across all samples (see Sec. B in Appendix for the full evaluation prompt).
To complement GPT-4o semantic evaluation and reduce potential stylistic bias from LLM-based judging, we additionally report objective tactile metrics including keyword-based F1 score and Top-5 Accuracy. F1 is computed based on tactile keyword overlap, while Top-5 Accuracy measures whether at least one predicted tactile attribute matches the ground-truth set.

\subsection{Implementation Details}
To validate \ours, we employ two high-performance, open-source sMLLMs: Qwen2.5-VL 3B~\cite{Qwen2.5-VL} and InternVL2.5-1B~\cite{chen2024expanding}.
These models represent the current state-of-the-art in compact multimodal intelligence, offering an ideal balance between sophisticated reasoning and deployment efficiency.
To minimize the computational footprint, we select a tactile encoder as a ViT-Tiny~\cite{dosovitskiy2021an} with $16 \times 16$ patches, initialized from ImageNet-pretrained~\cite{deng2009imagenet} weights.
This encoder has only 5.8M parameters.
A two-layer MLP projector with LayerNorm and GELU activation maps tactile features into the language embedding space. 
Both natural and tactile contact images are resized to $224\times224$. 
Following the Qwen2.5-VL formulation, tactile, visual, and textual tokens are concatenated into a single autoregressive sequence.
Training is performed using the AdamW optimizer~\cite{loshchilov2017decoupled} with weight decay set to 0.0. 
For \ours-3B, we train for 1 epoch with an initial learning rate of $2\times10^{-5}$, cosine decay scheduling and a warmup ratio of 0.05. For \ours-1B, we train for 2 epochs using the same optimization setup. All experiments are conducted on two NVIDIA RTX PRO 6000 Blackwell GPUs.
For mask-guided fine-tuning, the sparsity ratio is set to $s=60\%$.
For LoRA~\cite{lora} fine-tuning, we use rank $r=8$ with $\alpha=16$.
During inference, we use the following instruction \textit{``List 2-3 tactile attributes of the object shown in the image. Use single-word descriptors only. Output a comma-separated list with no additional text.''}, to generate tactile descriptions.

\subsection{Baselines}
We compare \ours with existing VTL approaches that extend LLMs including TVL-LLaMA~\cite{tvl} and UniTouch (Touch-LLM)~\cite{unitouch}.
Both methods employ 7B-parameter LLaMA models as the backbone, with UniTouch using LLaMA~\cite{llama} and TVL-LLaMA using LLaMA-2~\cite{touvron2023llama2}, and initialize the vision and tactile encoders using OpenCLIP~\cite{cherti2023reproducible}.
Among multiple tactile encoder scales (\eg, Tiny, Small, Base), we follow the ViT-Tiny setting. 
UniTouch adopts a larger ViT-L/14 encoder in its original design, and we follow this default configuration in our experiments.
Both approaches~\cite{tvl, unitouch} align tactile features with the language model and adapt the LLM using LoRA~\cite{lora}. 
By contrast, our \ours adopts modern vision-language models (VLMs), including Qwen2.5-VL-3B and InternVL2.5-1B, as the backbone. Since LLaMA-7B is a language-only backbone, it requires additional visual instruction tuning to perform zero-shot multimodal tasks. Therefore, to ensure a rigorous and fair comparison, we established a TVL-style LoRA baseline using the same Qwen2.5-VL-3B backbone adopted in \ours. 
This comparison isolates the effect of the adaptation strategy (LoRA vs.\ mask-guided fine-tuning) under identical initialization and training conditions.
We further compare against $\mathrm{(IA)}^3$~\cite{liu2022few} to assess a different PEFT paradigm that introduces learned scaling vectors instead of additive weight updates.

\subsection{Performance Comparison on VTL Tasks}
\begin{table*}[t]
\centering
\small
\caption{Overall performance comparison using LLM Judge to record scores in 10-point scale on three VTL benchmarks, including SSVTP~\cite{ssvtp}, TVL~\cite{tvl}, and TacQuad~\cite{feng2025anytouch}. \textbf{The best score} is bold, and \ul{the second} is underlined.}
\newcommand{\err}[1]{{\scriptsize $\pm$#1}}
\label{tab:main}
\resizebox{0.9\textwidth}{!}{
\setlength{\tabcolsep}{4pt}
\begin{tabular}{l l ccccc }
\toprule
{Method} & {Backbone} & SSVTP & TVL & TacQuad & Avg  \\
\midrule
Zero-shot & InternVL2.5-1B  & 3.61 & 3.54 & 3.45 & 3.53  \\

Zero-shot & Qwen2.5-VL-3B  & 3.74 & 3.40 & 3.57 & 3.57  \\
\midrule
UniTouch~\cite{unitouch} & LLaMA1-7B & 3.73 & 4.04 & 3.44 & 3.74  \\
TVL~\cite{tvl} & LLaMA2-7B & 5.27 & \ul{4.38} & 3.29 & 4.31  \\ 
TVL~\cite{tvl} & Qwen2.5-VL-3B & 4.98 & 4.29 & 4.22 & 4.50 \\
TVL-$\mathrm{(IA)}^3$~\cite{tvl,liu2022few} & Qwen2.5-VL-3B & 5.17 & 4.26 & 3.84 & 4.42 \\

\rowcolor{blueish!6} \textbf{\ours-1B} & InternVL2.5-1B & \textbf{5.69} &  4.34 & \textbf{5.02} &  \textbf{5.01}  \\
\rowcolor{blueish!6} \textbf{\ours-3B} & Qwen2.5-VL-3B &  \ul{5.48} & \textbf{4.39} &  \ul{4.86} & \ul{4.91}  \\
\bottomrule
\end{tabular}
}
\end{table*}
\cref{tab:main} compares performance on VTL tasks across three benchmarks.
We include zero-shot performance of pretrained VLMs without tactile adaptation as baselines.
Without tactile alignment, the zero-shot InternVL-1B and Qwen2.5-VL-3B models achieve average scores of only 3.53 and 3.57, respectively, indicating limited capability in reasoning about tactile properties.
Surprisingly, \ours-1B and -3B significantly outperform UniTouch, which uses a larger LLaMA1-7B backbone yet achieves only 3.74 on average. 
This highlights the effectiveness of the proposed mask-isolated adaptation even with smaller MLLMs.

\begin{wrapfigure}{r}{0.55\textwidth}
    \centering
    \vspace{-2em}
\includegraphics[width=1\linewidth]{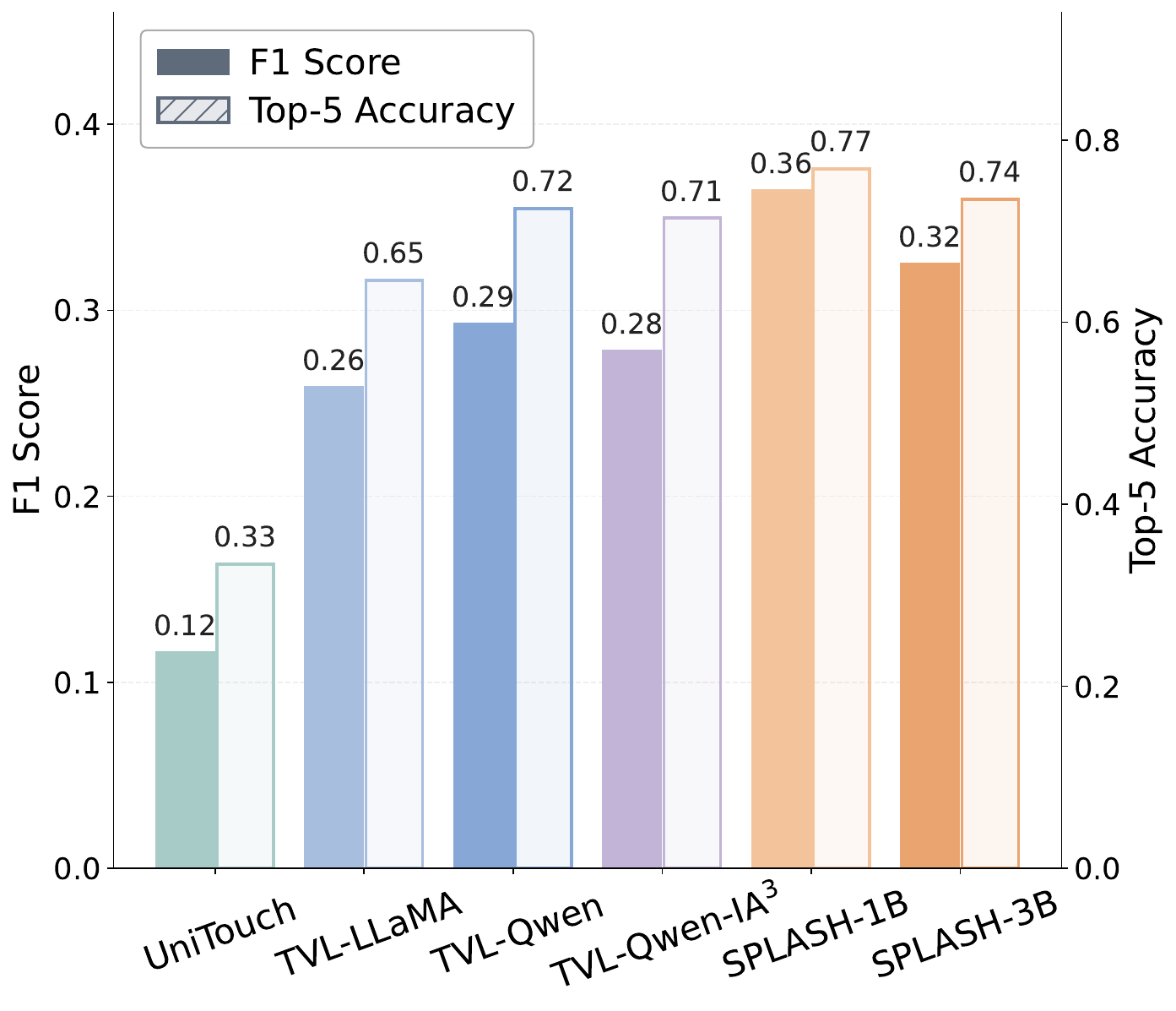}
    \caption{Quantitative comparison with objective metrics on SSVTP~\cite{ssvtp}, TVL~\cite{tvl}, and TacQuad~\cite{feng2025anytouch}. We report the averaged F1 score and Top-5 accuracy. \ours demonstrates superior performance in both F1 and Top-5 Accuracy.}
    \vspace{-1em}
    \label{fig:objective_metric_bar}
\end{wrapfigure}

Under the same Qwen2.5-VL-3B, \ours-3B achieves the best tactile-semantic performance among methods using the same backbone, reaching an average score of 4.91. 
In comparison, the strongest baseline, TVL~\cite{tvl}, with LoRA-based adaptation on Qwen2.5-VL-3B, achieves 4.50 on average. 
In particular, \ours-3B outperforms TVL 5.48 compared to 4.98 on SSVTP and 4.86 to 4.22 on TacQuad, suggesting strong visuo-tactile alignment and robustness under tactile distribution shifts. 
While $\mathrm{(IA)}^3$ shows competitive performance on relatively smaller VTL datasets like SSVTP and TVL, it suffers a significant performance drop on the unseen TacQuad, scoring only 3.84 (vs.\ 4.86 for \ours). This suggests that $\mathrm{(IA)}^3$ lacks the representational capacity required to generalize to real-world tactile dynamics. 

To verify that the effectiveness of our approach is not limited to a specific scale, we further employ \ours with a smaller InternVL-1B backbone. 
Despite its significantly smaller parameter size, \ours-1B achieves competitive tactile-semantic performance, reaching an average score of 5.01. 
These results suggest that the proposed mask-guided adaptation remains effective even for compact VLM backbones.

Beyond LLM-based evaluation, we quantitatively assess tactile attribute prediction with objective metrics. As shown in~\cref{fig:objective_metric_bar}, \ours consistently outperforms existing baselines in both F1 score and Top-5 Accuracy, indicating more precise tactile attribute generation and stronger robustness across diverse tactile vocabularies.

\subsection{Performance Comparison for Preservation of VL Capability}\
\begin{table*}[t]
\centering
\small
\caption{Overall performance comparison on five VL benchmarks, such as MMMUval~\cite{mmmu}, MathVista~\cite{mathvista}, MMEsum~\cite{mme}, MMBench-EN~\cite{mmbench}, and MMBench-CN~\cite{mmbench}. \textbf{The best score} is bold, and \ul{the second} is underlined.
}
\label{tab:main_vl}
\resizebox{\textwidth}{!}{
\setlength{\tabcolsep}{2pt}
\begin{tabular}{l l ccccc}
\toprule
Method & Backbone & MMMUval & MathVista & MMEsum & MMBench-EN & MMBench-CN\\
\midrule
Zero-shot & InternVL2.5-1B  & 40.9 & 43.2 & 1950 & 70.7 & 66.3\\
Zero-shot & Qwen2.5-VL-3B  & 53.1 & 62.3 & 2157 & 79.1 & 78.1 \\
\midrule
UniTouch~\cite{unitouch} & LLaMA1-7B &  26.6 & 9.0 & 27 & 8.5 & 8.6\\
TVL~\cite{tvl} & LLaMA2-7B &  26.0 & 11.6 & 582 & 8.6 & 5.5\\ 
TVL~\cite{tvl} & Qwen2.5-VL-3B &  50.0 & 52.8 & \ul{2168} & 76.8 & 76.5\\
TVL-$\mathrm{(IA)}^3$~\cite{tvl,liu2022few} & Qwen2.5-VL-3B & \ul{51.3} & \ul{63.0} & \bf 2216 & \bf 78.4 & \bf 78.9\\
\rowcolor{blueish!6} \textbf{\ours-1B} & InternVL2.5-1B  & 37.3 & 41.0 & 1786 & 70.2 & 60.4 \\
\rowcolor{blueish!6}
\textbf{\ours-3B} & Qwen2.5-VL-3B &  \bf 55.3 & \bf 65.3 & 2155 & \ul{78.0} & \ul{76.9} \\
\bottomrule
\end{tabular}
}
\end{table*}
\cref{tab:main_vl} evaluates VL capability retention after tactile adaptation on standard VL benchmarks. 
\ours-3B consistently outperforms the TVL baselines across all VL benchmarks.
In particular, \ours improves MMMUval from 50.0 to 55.3, MathVista from 52.8 to 65.3, and maintains comparable performance on MMEsum. 
Furthermore, \ours achieves higher scores on both MMBench-EN and MMBench-CN compared to TVL.
Compared with UniTouch~\cite{unitouch}, which relies on a larger LLaMA-7B backbone, \ours achieves substantially stronger VL reasoning performance across all benchmarks. 

We note that TVL-$\mathrm{(IA)}^3$~\cite{tvl, liu2022few} attains the highest scores on three benchmarks (\ie, MMEsum, MMBench-EN, and MMBench-CN), as its scaling-based adaptation perturbs the pretrained weights only minimally. 
This apparent advantage, however, comes at the expense of tactile learning: as shown in \cref{tab:main}, TVL-$\mathrm{(IA)}^3$ fails to generalize on TacQuad (3.84 vs.\ 4.86 for \ours-3B), revealing that its limited plasticity preserves VL ability only by under-fitting the tactile modality. In contrast, \ours attains the best scores on MMMUval and MathVista while remaining competitive on the remaining benchmarks, demonstrating that mask-guided isolation uniquely balances tactile acquisition with VL preservation -- rather than trading one for the other.
Notably, the performance of \ours remains comparable to the zero-shot Qwen2.5-VL-3B baseline, which achieves 53.1 on MMMUval and 62.3 on MathVista, confirming that tactile adaptation does not degrade the pretrained VL capability. We hypothesize that selectively updating VL-dormant subspaces acts as a form of pruning-induced regularization, which stabilizes the pretrained reasoning manifold during tactile adaptation~\cite{evci2020rigging}.

\subsection{Ablation Study}

\paragrapht{Sparsity ratio.}
\begin{table*}[t]
\centering
\small
\caption{The impact of sparsity ratio in \ours-3B. \textbf{The best score} is bold, and \ul{the second} is underlined. Based on the results, we establish 60\% as our base sparsity ratio ($s$=0.6).}
\label{tab:sparsity}
\resizebox{\textwidth}{!}{

\begin{tabular}{c cccc ccccc}
\toprule
\multirow{2}[2]{*}{Sparsity} & \multicolumn{4}{c}{Visuo-Tactile-Language} & \multicolumn{5}{c}{Vision-Language} \\ 
\cmidrule(lr){2-5} \cmidrule(lr){6-10}  & SSVTP  & TVL & TacQuad & Avg & MMMUval & MathVista & MMEsum & MMBench-EN & MMBench-CN\\
\midrule
100\%  & 5.02 & \ul{4.36} & 4.81 & 4.73 & 22.0 & 38.4 & 982 & 10.4 & 28.3 \\
80\% & 5.37 & 4.33 & \ul{4.85} & 4.85 & 52.6 & 57.9 & 2070 & 56.0 & 69.5 \\
\rowcolor{blueish!6} 60\% & \ul{5.48} & \bf 4.39 & \bf 4.86 & \bf 4.91 & \bf 55.3 & 65.3 & 2155 & 78.0 & 76.9\\
50\% & 5.10 & 4.32 & 4.74 & 4.72 & \ul{54.0} & \bf 65.8 & 2167 & \ul{78.3} & 77.7\\
40\% & \bf 5.60 & \ul{4.36} & 4.74 & \ul{4.90} & 52.6 & \ul{65.7} & \ul{2204} & \bf 78.6 & \ul{78.6} \\
30\% & 5.30 & 4.27 & 4.79 & 4.79 & 52.0 & 65.4 & \bf 2210 & 78.0 & \bf 78.8\\
\bottomrule
\end{tabular}
}
\end{table*}

\cref{tab:sparsity} presents an ablation study on the sparsity ratio used for mask-guided adaptation and its impact on both VTL and VL performance.
We observe that moderate sparsity levels provide the best trade-off between tactile-semantic performance and VL capability retention.
In particular, 60\% sparsity achieves the highest tactile-semantic average score of 4.91 while simultaneously obtaining the best VL reasoning performance on MMMUval and competitive results on MathVista and the MMBench benchmarks.
When the sparsity ratio becomes too large (\eg, 100\%), excessive parameter updates severely degrade VL reasoning performance, indicating catastrophic forgetting of pretrained knowledge. 
Conversely, when the sparsity ratio becomes too small (\eg, 30\%), excessive parameter freezing slightly limits tactile adaptation, leading to reduced tactile-semantic performance while VL capability remains largely preserved. These results highlight the importance of allocating an appropriate dormant subspace that balances tactile learning capacity with preservation of pretrained VL capabilities.
The analysis on the sparsity ratio for InternVL-1B is further reported in Sec. A.2 in Appendix.

\paragrapht{Calibration data.}
\begin{table*}[!t]
\centering
\small
\caption{Performance evaluation on calibration data choices, configurations, and sample sizes on \ours-3B. \ours demonstrates robust performance across varying setups and remains highly competitive even with unpaired noise data, indicating that it captures intrinsic architectural redundancies rather than overfitting to specific semantic alignments.}
\label{tab:abl_calibration}
\resizebox{\textwidth}{!}{
\begin{tabular}{l cccc ccccc}
\toprule
\multirow{2}[2]{*}{Calibration Setup} & \multicolumn{4}{c}{Visuo-Tactile-Language} & \multicolumn{5}{c}{Vision-Language} \\ 
\cmidrule(lr){2-5} \cmidrule(lr){6-10}  
& SSVTP  & TVL & TacQuad & Avg & MMMUval & MathVista & MMEsum & MMBench-EN & MMBench-CN\\
\midrule

\multicolumn{10}{l}{\textit{(a) Sample Sizes (Data = CC3M Paired)}} \\
\midrule
64 Paired          & 5.15 & 4.30 & 4.87 & 4.77 & \textbf{56.0} & 64.9 & \textbf{2168} & 77.7 & \textbf{77.4} \\
128 Paired (Ours)  & \textbf{5.48} & \textbf{4.39} & 4.86 & \textbf{4.91} & 55.3 & 65.3 & 2155 & \textbf{78.0} & 76.9 \\
256 Paired         & 5.28 & 4.35 & 4.86 & 4.83 & \textbf{56.0} & \textbf{65.6} & 2146 & 77.7 & 77.2 \\
\midrule
\multicolumn{10}{l}{\textit{(b) Data Choices \& Configurations (Sample Size = 128)}} \\
\midrule
CC3M Paired (Ours) & \textbf{5.48} & \textbf{4.39} & 4.86 & \textbf{4.91} & 55.3 & \textbf{65.3} & 2155 & \textbf{78.0} & 76.9 \\
TVL Paired         & 4.98 & 4.35 & \textbf{4.93} & 4.75 & \textbf{58.0} & 62.9 & 2156 & 76.5 & 76.8 \\
Unpaired (Noise)   & 5.45 & 4.31 & 4.89 & 4.88 & 55.3 & 64.9 & \textbf{2166} & 77.7 & \textbf{77.4} \\

\bottomrule
\end{tabular}
}
\end{table*}
\cref{tab:abl_calibration} evaluates the effectiveness of different calibration datasets used to construct the dormant subspace mask for \ours-3B. 
We first ablate the calibration sample size. The proposed method remains highly stable across 64, 128, and 256 samples, showing only marginal variations across benchmarks. While all configurations achieve competitive performance, we select 128 samples as the default setting since it provides the optimal balance among VL reasoning performance, stability, and calibration efficiency.
In addition, using either CC3M or TVL for the calibration dataset yields comparable performance on both VTL and VL benchmarks (\cref{tab:abl_calibration}(b)). 
The VTL average score changes only slightly from 4.91 to 4.75, while the VL benchmark results remain largely stable.
These observations suggest that the proposed visual-relative importance metric is robust to the choice of calibration corpus.
Most surprisingly, we further assess an extreme setup where vision-language semantically calibrated data is unavailable (\ie, using unpaired random noise images with semantically meaningless dummy text prompts). 
The performance on the `unpaired' setting in \cref{tab:abl_calibration}(b) shows this calibration-free variant still achieves highly competitive performance. 
Consequently, \ours successfully captures the intrinsic architectural redundancies of modern MLLMs, rather than merely overfitting to dataset-specific activation patterns.

\begin{figure}[p]
    \centering
    \begin{subfigure}{0.99\linewidth}
        \centering{
        \includegraphics[width=\linewidth]{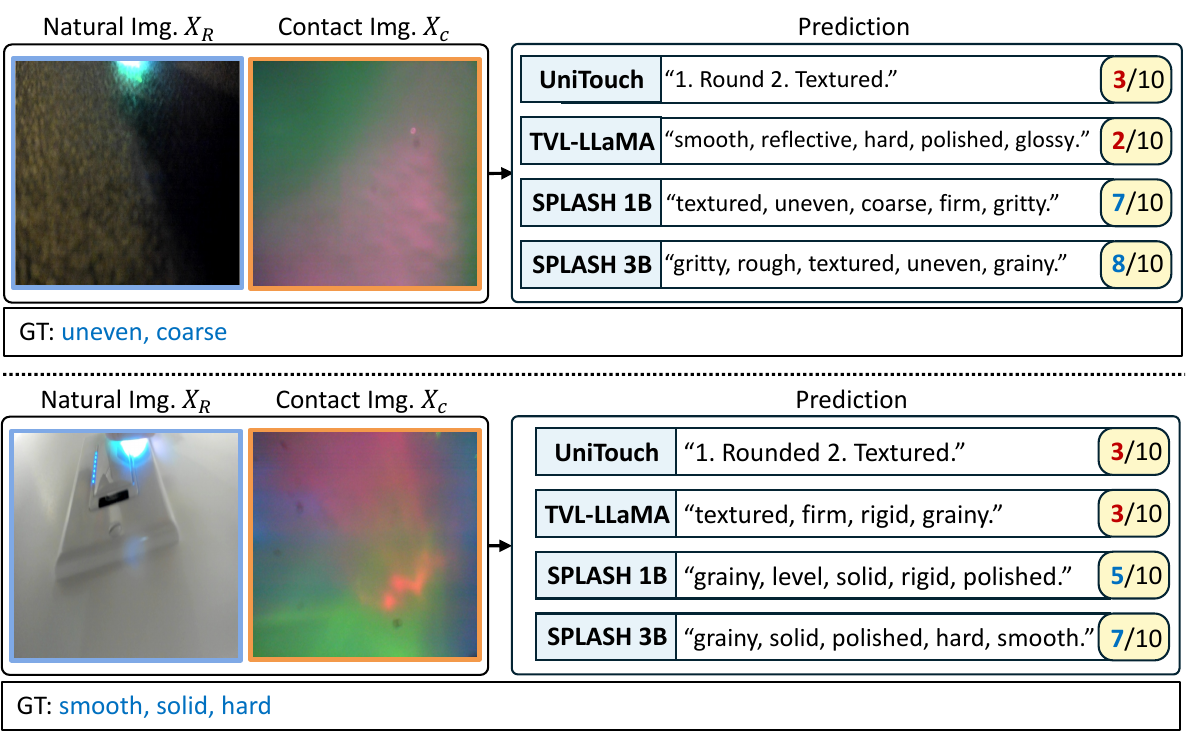}\vspace{-1em}
        }
        \caption{TVL}
    \end{subfigure} \\
    \begin{subfigure}{0.99\linewidth}
        \centering{
        \includegraphics[width=\linewidth]{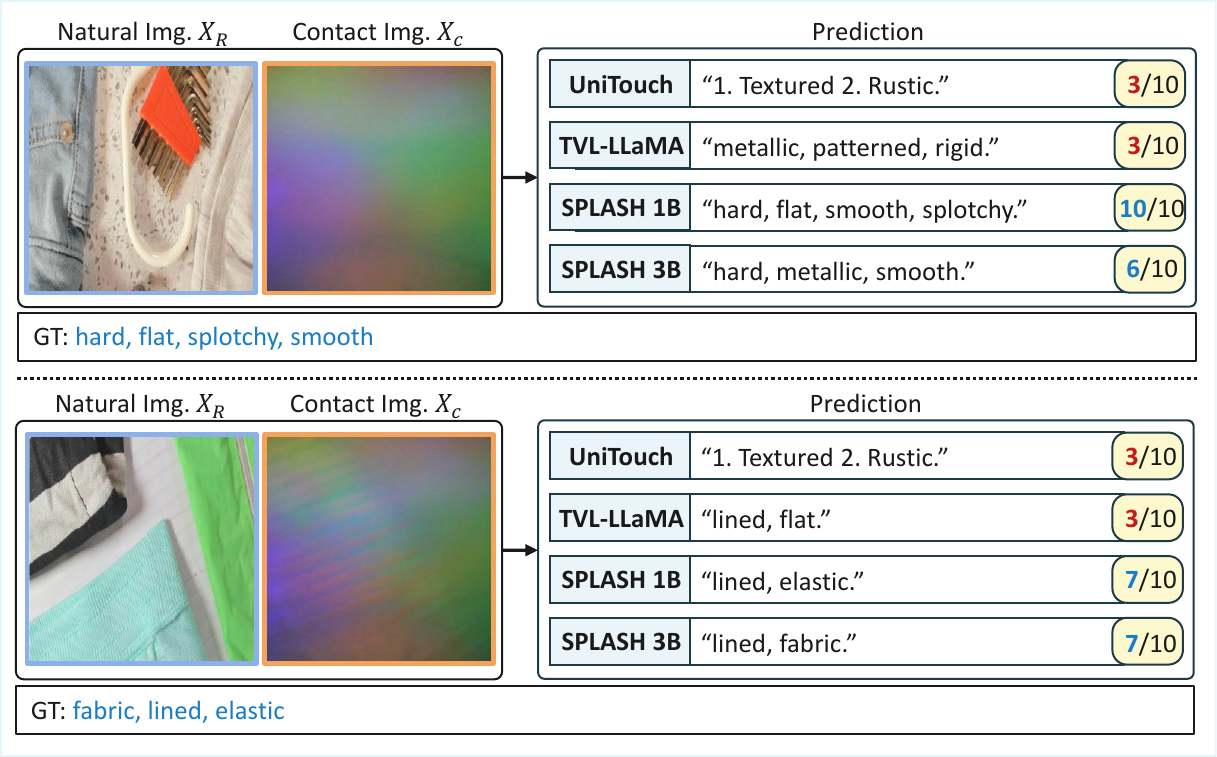}\vspace{-1.2em}
        }
        \caption{SSVTP}
    \end{subfigure}
    \caption{Qualitative comparisons on TVL~\cite{tvl} and SSVTP~\cite{ssvtp}. \ours demonstrates the robustness with higher accuracy score than TVL-LLaMA7B~\cite{tvl} and UniTouch~\cite{unitouch}. Especially, \ours-1B highlights the best scores in examples on SSVTP, even at the 1B-parameter scale.}
    \label{fig:qualitative1}
\end{figure}

\paragrapht{Tactile frontend initialization.}
\begin{table}[t]
\centering
\caption{Comparison using the same tactile-pretrained frontend initialization adopted in TVL.}
\scriptsize
\setlength{\tabcolsep}{1pt}
\resizebox{\columnwidth}{!}{
\begin{tabular}{c cccc ccccc}
\toprule
\multirow{2}[2]{*}{Method} & \multicolumn{4}{c}{Visuo-Tactile-Language} & \multicolumn{5}{c}{Vision-Language} \\ 
\cmidrule(lr){2-5} \cmidrule(lr){6-10}
& SSVTP & TVL & TacQuad & Avg & MMMUval & MathVista & MMEsum & MMBench-EN & MMBench-CN \\
\midrule
TVL(Qwen2.5-VL-3B) & 4.98 & 4.29 & 4.22 & 4.50 & 50.0 & 52.8 & 2168 & 76.8 & 76.5 \\
\textbf{\ours-3B} & \textbf{5.10} & \textbf{4.35} & \textbf{4.88} & \textbf{4.78} & \textbf{56.0} & \textbf{65.3} & \textbf{2195} & \textbf{78.7} & \textbf{76.7} \\
\bottomrule
\end{tabular}
}
\label{tab:abl_frontend_init}

\end{table}
To isolate the effect of tactile frontend initialization, we additionally compare \ours-3B and TVL using the same tactile-pretrained encoder initialization adopted in TVL. 
In \cref{tab:abl_frontend_init}, \ours consistently outperforms TVL across both VTL and VL benchmarks, indicating that the performance gain mainly stems from the proposed mask-guided adaptation strategy rather than the frontend initialization itself.

\subsection{Qualitative Results}
\begin{figure}[t]
    \centering
    \includegraphics[width=\linewidth]{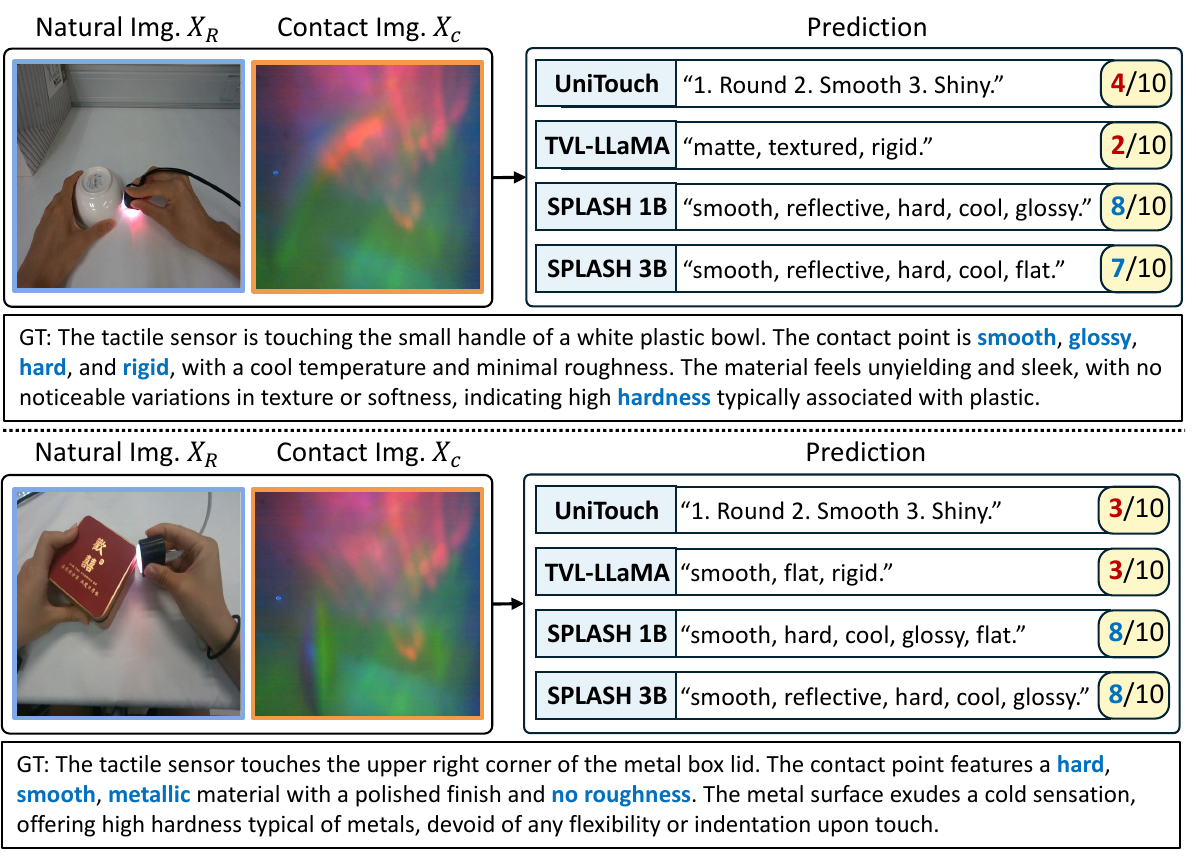}
    \caption{Qualitative comparisons on TacQuad~\cite{feng2025anytouch}.
    Both \ours-1B and -3B generate more diverse predictions than baselines, achieving higher accuracy scores by an LLM judge.
    }
    \label{fig:qualitative2}
\end{figure}

\cref{fig:qualitative1,fig:qualitative2} present qualitative comparisons across the VTL benchmarks, including TVL, SSVTP, and TacQuad.
Given paired RGB and tactile contact images, models are asked to generate tactile attribute descriptions.
Across all benchmarks, \ours produces closely matched descriptions of the ground-truth tactile properties. TVL-LLaMA and UniTouch often generate visually plausible but tactually inconsistent attributes (\eg, smooth or reflective for coarse surfaces) or generic attributes unrelated to the underlying tactile properties. These mismatches lead to lower semantic similarity scores in the GPT-4o evaluation, whereas our models consistently achieve higher scores across the examples. 
Overall, the results suggest that baselines rely more heavily on visual appearance cues when tactile grounding is weak, whereas \ours more effectively integrates tactile observations to infer physically consistent surface properties.
More qualitative examples are provided in Sec. A.8 in Appendix.

\subsection{Limitation and Future Work}
While \ours demonstrates a robust capability for non-destructive tactile adaptation, the critical subspace is static after the offline step to split parameters.
In practical robotic manipulation scenarios, however, activation patterns may dynamically change depending on interaction state, contact condition, or task progression. Thus, a static dormant subspace may not fully capture temporally varying parameter utilization during long-horizon visuo-tactile reasoning.
Future work explores dynamically adaptive masking strategies that update dormant parameter allocation according to task-dependent activation statistics during embodied interaction.
Moreover, our current experiments focus on a single DIGIT tactile sensor setup, and cross-sensor generalization remains an important direction for future investigation. Nevertheless, as \ours is designed in a frontend-agnostic manner, it can be naturally extended to incorporate alternative tactile encoders. 

\section{Conclusion}
We present \textbf{\ours}, a novel mask-isolated tactile alignment framework that expands the sensory capability of small MLLMs without catastrophic forgetting.
\ours identifies and isolates a dormant parameter via a visual-relative importance metric.
During training, only the dormant subspace is updated simultaneously with the tactile front-end, while the critical parameters responsible for foundational VL reasoning stay anchored.
Extensive experiments across diverse VTL and VL benchmarks demonstrate that \ours effectively ``wake up'' underutilized parameters to internalize complex contact dynamics, achieving state-of-the-art tactile reasoning while fully preserving general-purpose ability.
With its unified single-stage training and zero added inference cost, we believe \ours offers a practical path toward high-level tactile intelligence for resource-constrained edge robots.

\section*{Acknowledgements}
This work was supported by the National Research Foundation of Korea(NRF) grant funded by the Korea government(MSIT) (RS-2025-16065706).

\bibliographystyle{splncs04}
\bibliography{main}

\newpage
\appendix
\setcounter{table}{0}
\setcounter{figure}{0}
\renewcommand\thetable{\thesection\arabic{table}}
\renewcommand\thefigure{\thesection\arabic{figure}}

\section*{\Large Appendix}
\section{Additional Discussion}

\subsection{Vision Forgetting Problem}
\begin{figure}[h]
    \centering
    \includegraphics[width=\textwidth]{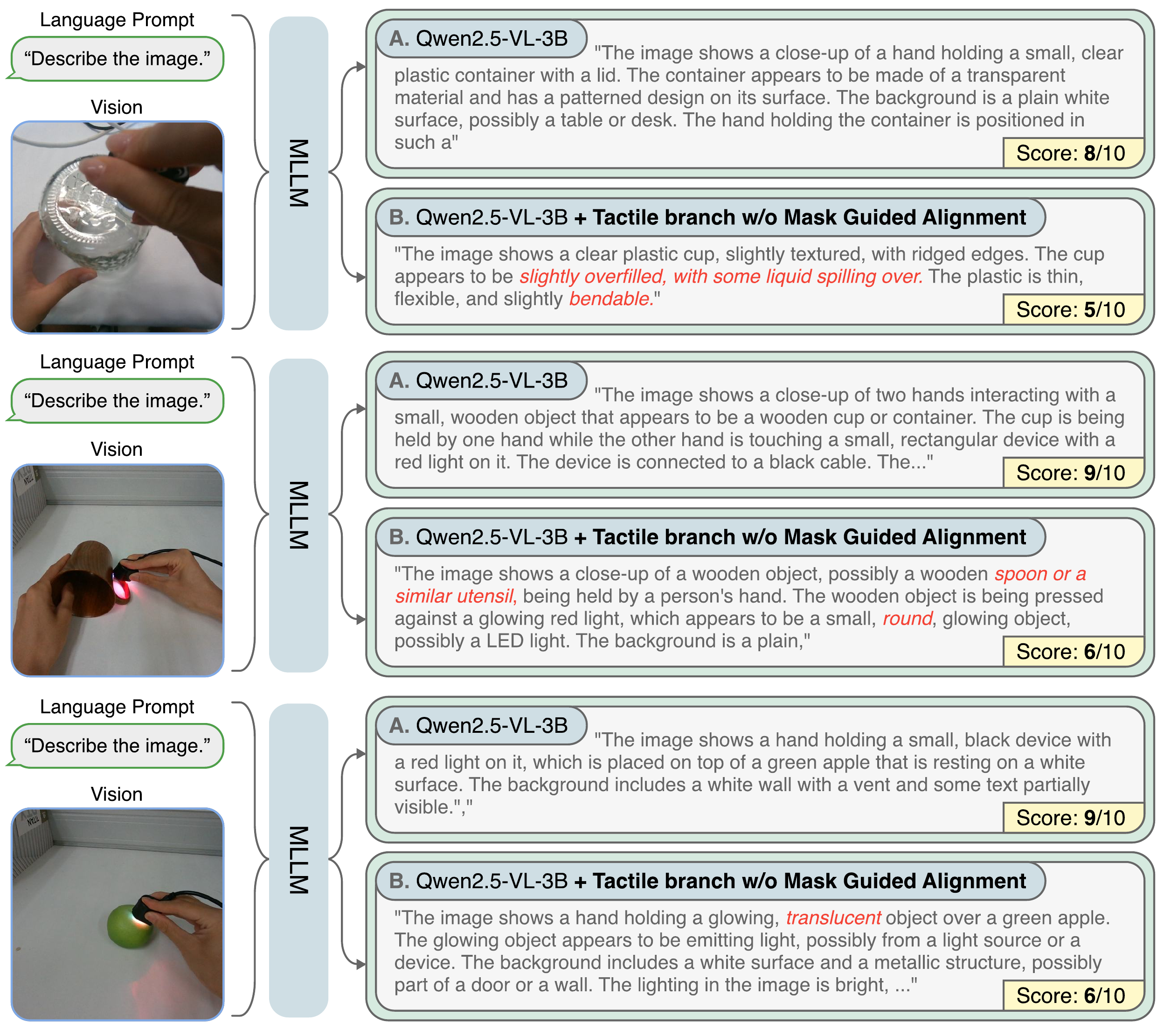}
    \caption{
        {More examples of the catastrophic forgetting problem in tactile alignment for MLLMs.} 
    }
    \label{fig:suppl_forgetting}
\end{figure}

Our motivation relies on the catastrophic forgetting problem, especially in the visual domain.
To illustrate this issue, we present additional qualitative examples in \cref{fig:suppl_forgetting}. 
Although the original Qwen2.5-VL correctly recognizes visual objects and scenes, the previous tactile-aligned model (\ie, TVL) frequently produces inaccurate descriptions or hallucinated attributes that are not supported by the visual input. 

\subsection{Sparsity Analysis for SPLASH-1B}\label{app:1bsparsity}
\begin{table*}[t]
\centering
\small
\caption{The impact of sparsity ratio in \ours-1B. \textbf{The best score} is bold, and \ul{the second} is underlined. 
}
\label{tab:abl_1b}
\resizebox{\textwidth}{!}{
\begin{tabular}{c cccc ccccc}
\toprule
\multirow{2}[2]{*}{Sparsity} & \multicolumn{4}{c}{Visuo-Tactile-Language} & \multicolumn{5}{c}{Vision-Language} \\ 
\cmidrule(lr){2-5} \cmidrule(lr){6-10}  & SSVTP  & TVL & TacQuad & Avg & MMMUval & MathVista & MMEsum & MMBench-EN & MMBench-CN\\
\midrule
InternVL2.5-1B  & 3.61 & 3.54 & 3.45 & 3.53 & 40.9 & 43.2 & 1950.5 & 70.7 & 66.3 \\
\midrule
100\%  & 5.89 & 4.30 & \bf 5.07 & 5.09 & 1.3 & 19.0 & 48 & 2.1 & 4.4 \\
80\%  & \bf 6.15 & \ul{4.32} & 4.93 & \ul{5.13} & 34.6 & 34.0 & 1474 & 62.3 & 55.0 \\
\rowcolor{blueish!6} 60\%  & 5.69 & 4.34 & 5.02 & 5.01 & \ul{37.3} & 41.0 & 1786 & 70.2 & 60.4 \\
50\% & \ul{6.13} & 4.30 & 5.02 & \bf 5.15 & 36.0 & 42.6 & 1782 & 69.8 & 61.6 \\
40\% & 5.52 & \bf 4.33 & \ul{5.04} & 4.96 & 36.6 & \bf 44.1 & \ul{1839} & \bf 71.2 & \ul{62.7} \\
30\% & 5.58 & \bf 4.33 & 4.99 & 4.97 & \bf 42.6 & \ul{43.4} & \bf 1930 & \ul{70.8} & \bf 63.4 \\
\bottomrule
\end{tabular}
}
\end{table*}

We further analyze the effect of sparsity on \ours-1B.
As shown in \cref{tab:abl_1b}, a sparsity ratio of $60\%$ provides a favorable balance between VTL and VL performance for \ours-1B. 
The VL performance of \ours-1B varies more noticeably between sparsity levels, whereas VL benchmarks for \ours-3B remain relatively stable between sparsity ratios of 30\% to 60\%.
While a sparsity ratio of $60\%$ with \ours-3B provides a good balance between VTL and VL performance, applying the same sparsity ratio to the smaller 1B backbone leads to slightly lower VL performance than lower sparsity levels (30\%-40\%).
Therefore, the smaller capacity of 1B backbone makes it more sensitive to the ratio of masking during tactile adaptation.

\subsection{Adaptive Sparsity Allocation}

Prior work demonstrates that different transformer layers exhibit varying levels of redundancy and sensitivity during weight pruning~\cite{yin2024outlier}. Motivated by this observation, we further explore an adaptive sparsity allocation strategy that optimizes layer-wise sparsity while maintaining a fixed global sparsity budget.

Specifically, following the OWL~\cite{yin2024outlier} framework, we first estimate the Layer-wise Outlier Distribution (LOD) using activation-weight importance statistics computed over a small vision-language calibration set. Layers containing more activation outliers are regarded as more important for preserving pretrained representations. We therefore allocate lower sparsity to layers with higher LOD scores and higher sparsity to layers with lower LOD scores. To avoid excessive imbalance across transformer blocks, the layer-wise sparsity ratios are constrained within a bounded range around the target global sparsity ratio and then globally rescaled to preserve the exact overall sparsity budget.

\begin{figure}[t]
    \centering
    \includegraphics[width=\textwidth]{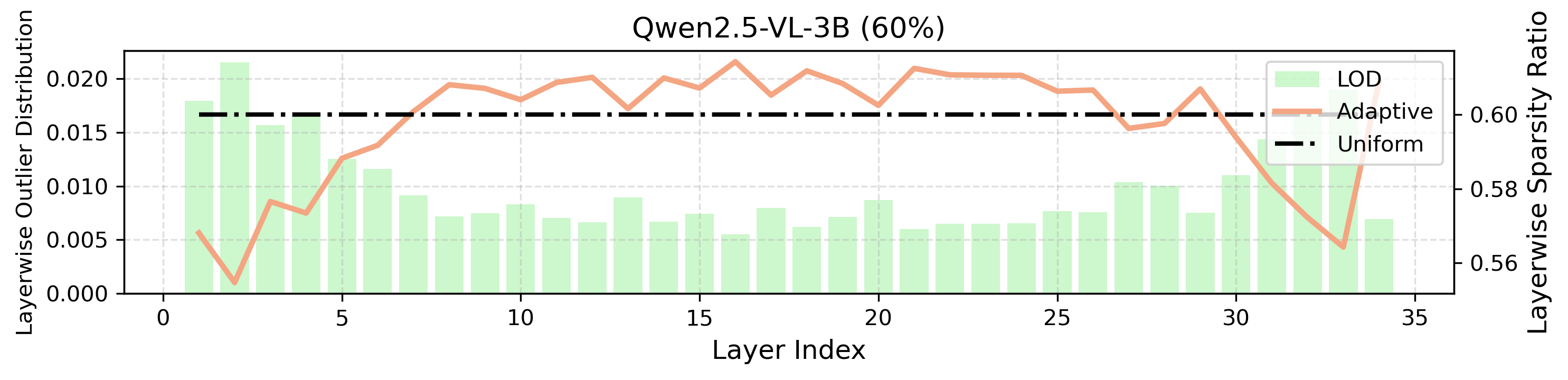}
    \caption{
    OWL-inspired adaptive sparsity allocation under a global 60\% sparsity budget. Green bars denote the Layer-wise Outlier Distribution (LOD), while the orange curve represents the resulting adaptive sparsity ratio assigned to each transformer layer.
    }
    \label{fig:owl_adaptive_sparsity}
\end{figure}

\begin{table}[t]
\centering
\caption{Comparison between adaptive and uniform sparsity allocation strategies under the same global sparsity budget ($s=60\%$).}
\label{tab:adaptive_sparse}

\scriptsize
\setlength{\tabcolsep}{2pt}
\resizebox{\columnwidth}{!}{
\begin{tabular}{c cccc ccccc}
\toprule

\multirow{2}[2]{*}{Strategy} 
& \multicolumn{4}{c}{Visuo-Tactile-Language} 
& \multicolumn{5}{c}{Vision-Language} \\ 

\cmidrule(lr){2-5} 
\cmidrule(lr){6-10}

& SSVTP & TVL & TacQuad & Avg 
& MMMUval & MathVista & MMEsum & MMBench-EN & MMBench-CN \\

\midrule

Adaptive 
& 5.26 & 4.37 & 4.85 & 4.83 
& 54.7 & 64.8 & \textbf{2160} & 77.3 & \textbf{77.1} \\

Uniform 
& \textbf{5.48} & \textbf{4.39} & \textbf{4.86} & \textbf{4.91} 
& \textbf{55.3} & \textbf{65.3} & 2155 & \textbf{78.0} & 76.9 \\

\bottomrule
\end{tabular}
}

\end{table}

\cref{fig:owl_adaptive_sparsity} visualizes the resulting layer-wise sparsity distribution for \ours-3B under a global 60\% sparsity setting.
\cref{tab:adaptive_sparse} compares the adaptive sparsity strategy with the default uniform sparsity allocation. While an adaptive sparsity strategy yields marginal gains, uniform sparsity consistently dominates VTL metrics and overall performance. Therefore, uniform sparsity ensures balanced tactile flow in dormant spaces.

\subsection{Dormant Subspace Importance Scores}
To better understand the internal mechanics of \ours during tactile alignment, we analyze the parameter shifts within the trainable dormant subspace (\ie, $\mathbf{M}_{ij}=1$) between the original pretrained Qwen2.5-VL 3B and our \ours-3B ($s=60\%$). We evaluate both models on a subset of TVL. 
Following our masking strategy, we extract the weights from the linear layers within the intermediate transformer blocks, excluding the first and last blocks. We uniformly sample 15 million dormant parameters across these layers.

\begin{figure}[t]
    \centering
    \includegraphics[width=\textwidth]{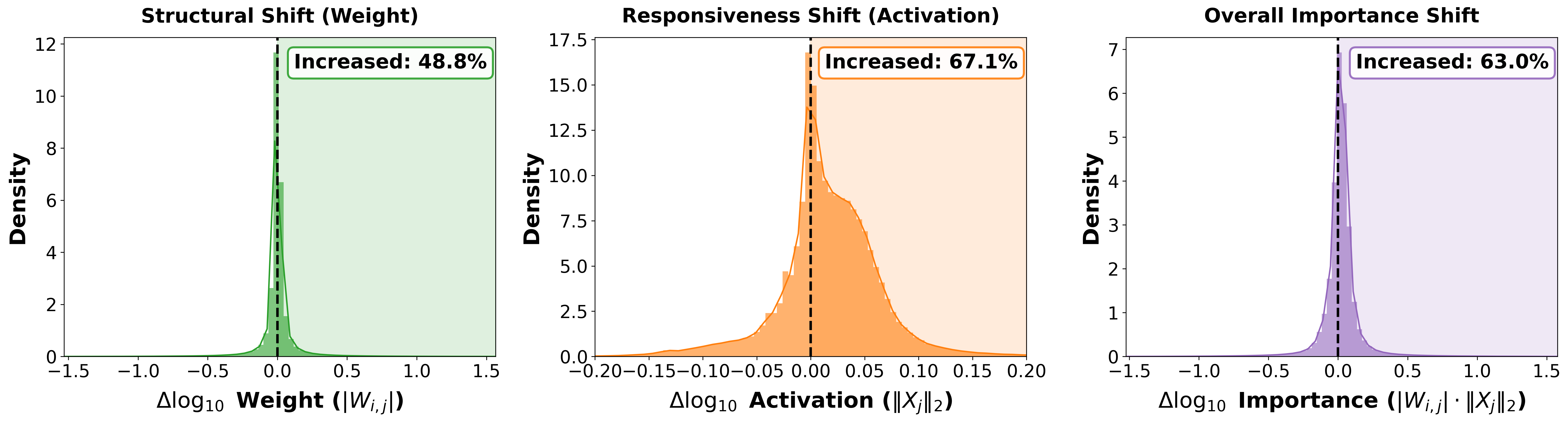}
    \caption{
        {Decomposed parameter shifts in the trainable dormant subspace.} 
        The plots show the $\Delta \log_{10}$ distributions (\ours minus baseline) for {(Left)} weight magnitude $|\weight_{i,j}|$, {(Middle)} activation $\ell_2$-norm $\|\act_j\|_2$, and {(Right)} overall importance score $\score_{i,j}$. 
        The prominent rightward shift in activation confirms its primary role in accommodating the newly introduced tactile modality.
    }
    \label{fig:parameter_shift}
\end{figure}

We decompose the shift in the visual-relative importance score, $\score_{i,j} = |\weight_{i,j}| \cdot \| \act_j \|_2$, into its two constituent factors: the structural shift in weight magnitude ($|\weight_{i,j}|$) and the responsiveness shift in the activation $\ell_2$-norm ($\|\act_j\|_2$).  \cref{fig:parameter_shift} illustrates the $\Delta \log_{10}$ distributions of these components.
Interestingly, 48.8\% of the dormant weights exhibit an increase in magnitude. 
Conversely, the responsiveness shift (\cref{fig:parameter_shift}, middle) demonstrates a significant majority (67.1\%) of the corresponding activations exhibit a strong positive shift in their $\ell_2$-norm. This strongly suggests that the dormant pathway becomes highly sensitized and explicitly responsive to the newly introduced tactile inputs.
Consequently, 63.0\% of the dormant weights show increased scores (\cref{fig:parameter_shift}, right). These findings empirically validate our selective plasticity design: \ours successfully accommodates new modalities by dynamically amplifying the input-driven activations along the selectively trainable dormant pathways.

\subsection{Freezing Boundary Layers}
Following prior work~\cite{gromov2025unreasonable}, we freeze the first and last transformer layers during visuo-tactile adaptation, as these layers are known to be particularly sensitive to distribution shifts and closely tied to preserving pretrained representations.
Empirically, we observe that unfreezing both boundary layers reduces the average VTL score by 0.18 points, indicating that updating these layers can negatively affect the balance between tactile adaptation and visual-language preservation.
Therefore, we keep the boundary layers frozen throughout all experiments for stable adaptation.

\subsection{Inference cost}
\begin{table}[t]
\centering
\small
\caption{Deployment cost comparison on NVIDIA RTX PRO 6000 Blackwell using a fixed TVL sample. P99 latency denotes the 99th percentile latency measured over the steady-state inference runs.}
\label{tab:deployment}
\setlength{\tabcolsep}{4pt}
\begin{tabular}{ll cc}
\toprule
Model & Backbone  
& P99 Latency (ms)$\downarrow$
& Control Hz$\uparrow$
\\
\midrule
TVL & LLaMA-7B &  202.37 & 4.97 \\
TVL, \ours & Qwen2.5-VL-3B & \textbf{134.55} & \textbf{7.50} \\
\bottomrule
\end{tabular}
\end{table}
We measure end-to-end inference latency of all methods on a fixed TVL sample using a single NVIDIA RTX PRO 6000 Blackwell GPU with batch size 1. Each model is evaluated under identical input configurations and prompts, generating up to 6 output tokens. To ensure stable measurements, we perform 50 warm-up iterations followed by 200 inference runs, where the first 50 runs are discarded to account for initialization overhead. For the Qwen2.5-VL-based TVL baseline, the latency is measured after merging the LoRA weights into the backbone model. 
As shown in \cref{tab:deployment}, both TVL-Qwen2.5-VL and \ours achieve a P99 latency of 134.55 ms and a control frequency of 7.50 Hz.
Compared with the LLaMA-7B-based TVL baseline (202.37 ms, 4.97 Hz), both methods exhibit substantially lower latency due to the smaller Qwen2.5-VL-3B backbone used during inference. These results demonstrate that tactile reasoning can be seamlessly integrated into compact VLMs without compromising inference efficiency, making \ours highly suitable for real-time robotic perception loops. 

\begin{figure}[t]
    \centering
    \includegraphics[width=\linewidth, page=1]{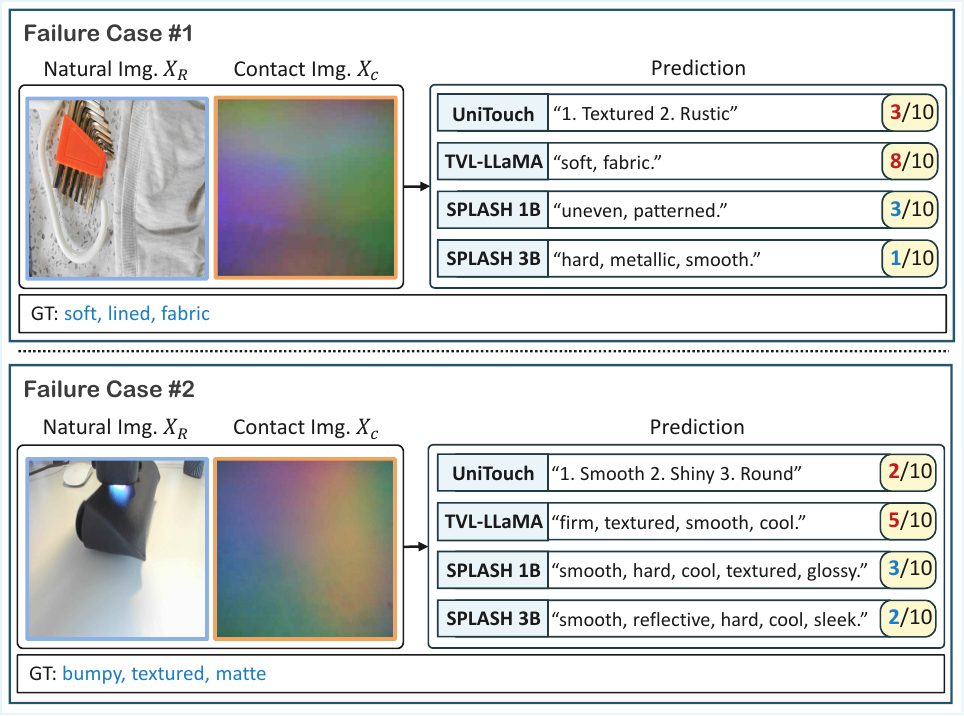}
    \caption{
        {Qualitative failure cases in visuo-tactile perception.} 
        (Top) Material misclassification on SSVTP, where a soft fabric surface is incorrectly predicted as a rigid metallic material. 
        (Bottom) Micro-texture misinterpretation on TVL, where a bumpy matte surface is predicted as smooth and glossy. 
    }
    \label{fig:failure_cases}
\end{figure}

\subsection{Failure Case Analysis}

\noindent\textbf{Case 1: Material misclassification under tactile perceptual aliasing.}
In the top of \cref{fig:failure_cases}, the first case originates from SSVTP, where our proposed models fail to correctly identify the textile material properties. 
We argue that tactile perceptual aliasing~\cite{lloyd2021probabilistic} introduces inherent ambiguity, where, when a fabric surface experiences minimal deformation upon contact, the resulting tactile imprint often lacks distinctive structural patterns and becomes geometrically similar to that of a smooth rigid surface.
Consequently, the repetitive linear texture of the fabric appears visually similar to brushed metallic patterns, causing \ours-3B to generate rigid descriptors like ``metallic'' and ``smooth'', and \ours-1B's prediction (``uneven, patterned'').

\noindent\textbf{Case 2: Micro-texture misinterpretation and Visual Dominance.}
In the bottom of \cref{fig:failure_cases}, all competitive models, including \ours, misinterpret the underlying surface micro-geometry.
This error indicates a failure to properly integrate tactile signals that encode fine-grained surface geometry, which can be effectively captured by high-resolution tactile sensors such as GelSight at micron-level resolution~\cite{yuan2017gelsight}. 

\subsection{Additional Qualitative Results}\label{app:morequal}
\begin{figure}[t]
    \centering
    \includegraphics[width=\linewidth, page=1]{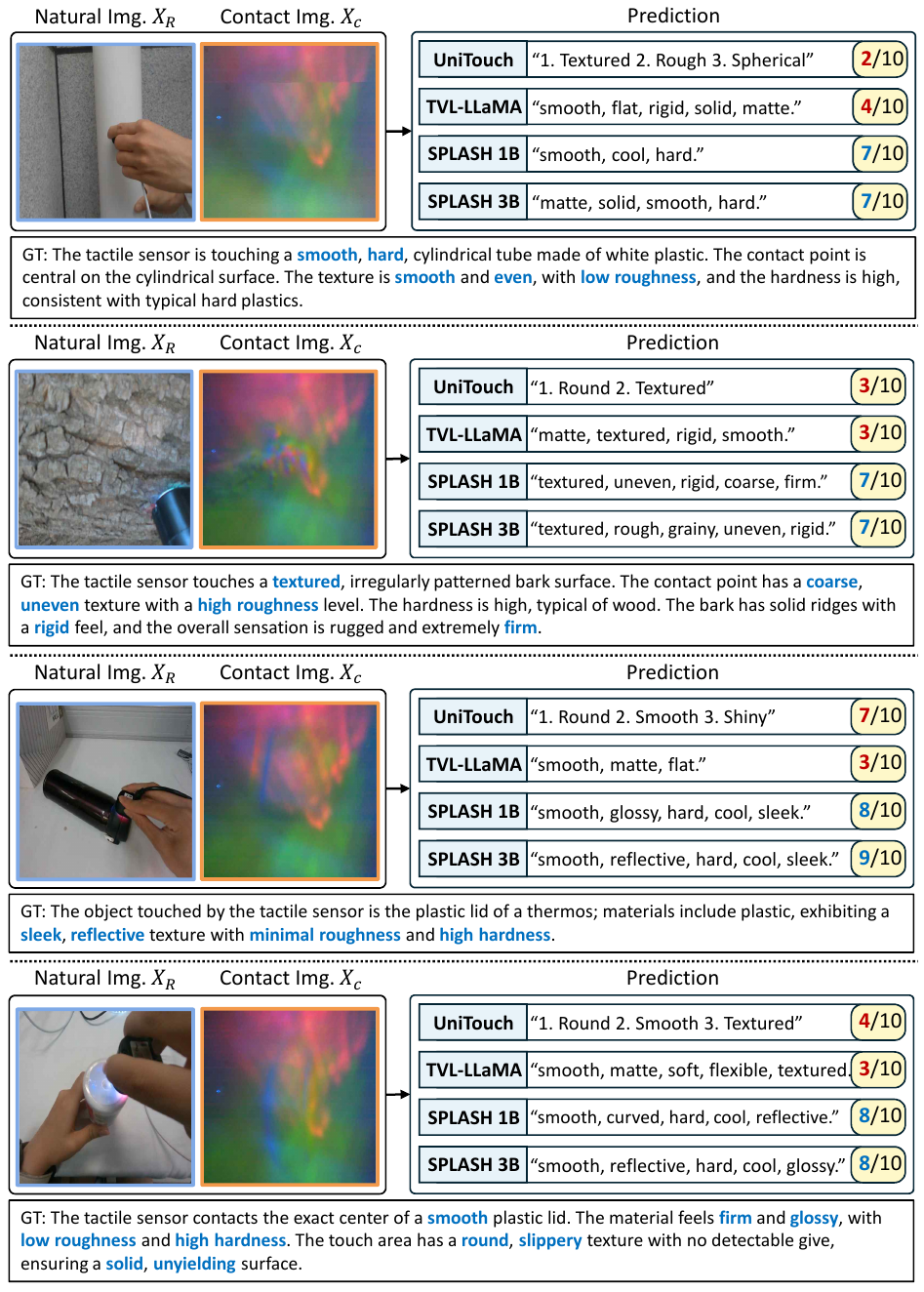}
    \caption{
        {Additional qualitative comparisons on TacQuad.} 
        Each example displays the natural RGB image $X_R$, the tactile contact image $X_C$, and the predicted texts from the baselines and our models alongside their evaluation scores (out of 10). 
    }
    \label{fig:qual_add1}
\end{figure}

\begin{figure}[t]
    \centering
    \includegraphics[width=\linewidth, page=2]{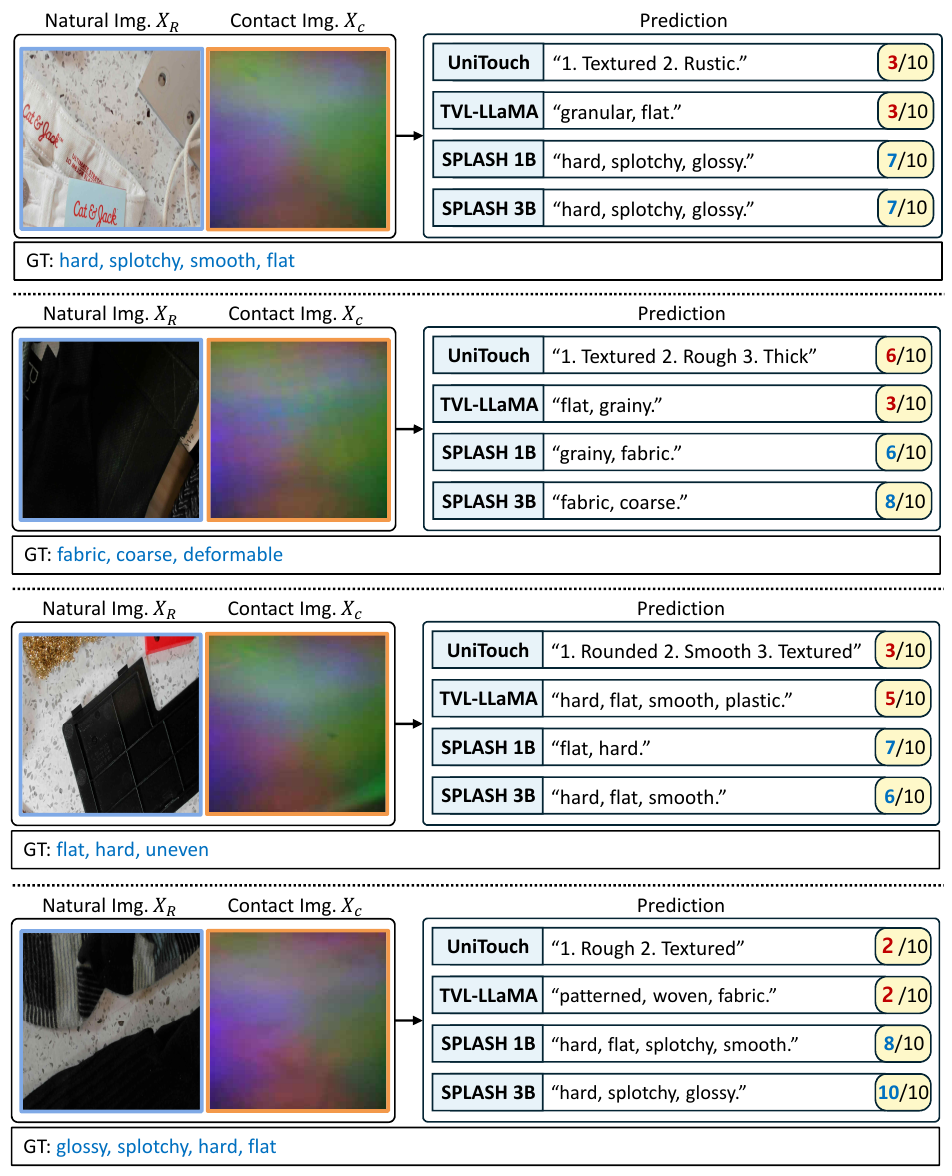}
    \caption{
        {Additional qualitative comparisons on SSVTP.} 
    }
    \label{fig:qual_add2}
\end{figure}

\begin{figure}[t]
    \centering
    \includegraphics[width=\linewidth, page=3]{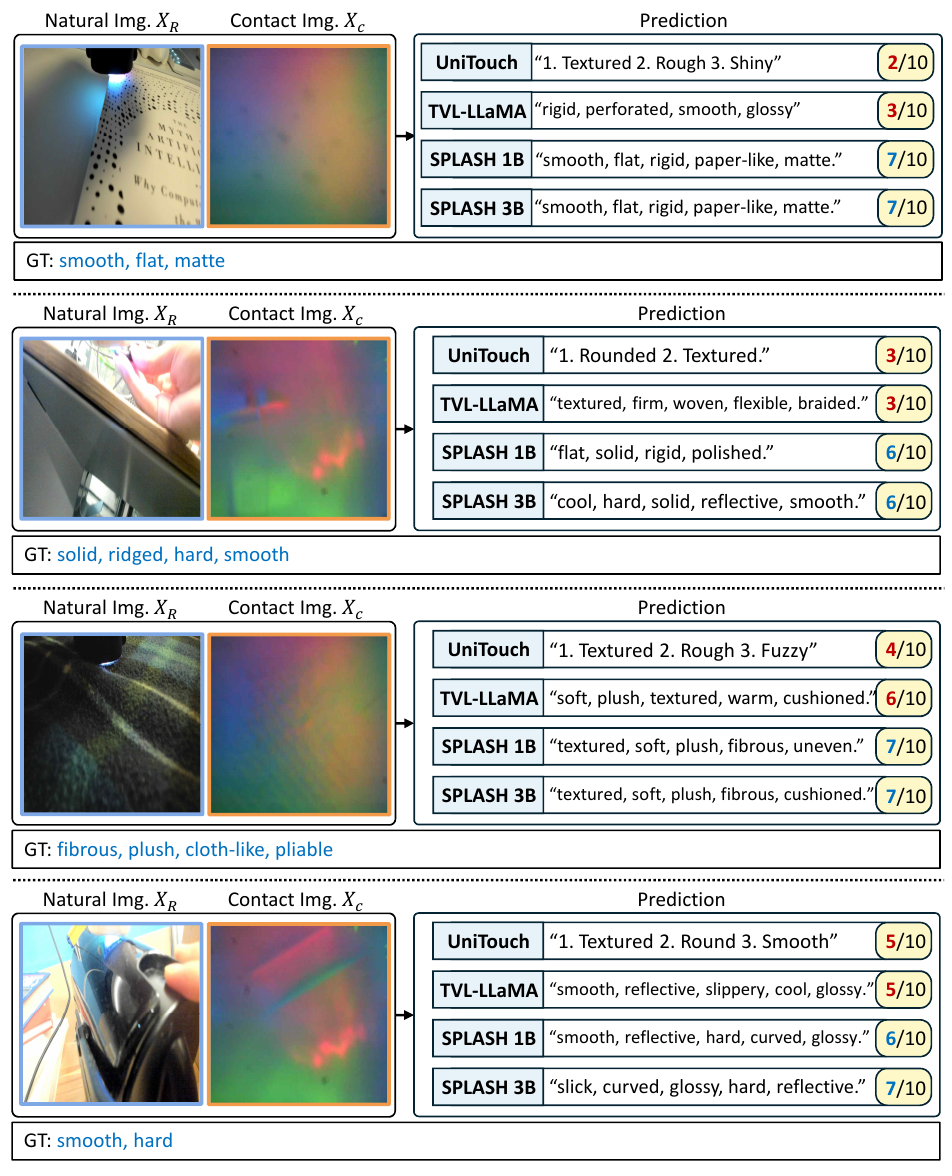}
    \caption{
        Additional qualitative comparisons on TVL.
    }
    \label{fig:qual_add3}
\end{figure}

We provide additional qualitative examples across the TacQuad, SSVTP, and TVL benchmarks in \cref{fig:qual_add1}, \cref{fig:qual_add2}, and \cref{fig:qual_add3}. These examples further illustrate that our proposed models produce more diverse and detailed tactile descriptions compared to the baseline methods.

\section{LLM Judge Prompt}
\label{sec:eval_protocol}

Following TVL, we use GPT-4o as an LLM judge to evaluate tactile descriptions using the following prompt on VTL benchmarks.

\begin{tcolorbox}[colback=gray!5!white, colframe=gray!75!black, title=VTL Benchmark Evaluation Prompt]
[User Question]: \{prompt\}

[Assistant Response]: \{assistant\_response\}

[Correct Response]: \{correct\_response\}

We would like to request your feedback on the performance of an AI assistant in response to the user question displayed above.
The user asks the question on observing an image. The assistant's response is followed by the correct response.

Please evaluate the assistant's response based on how closely it matches the correct response which describes tactile feelings. Please compare only the semantics of the answers. DO NOT consider grammatical errors in scoring the assistant. The assistant receives an overall score on a scale of 1 to 10, where a higher score indicates better overall performance.

Please first output a single line containing only one value indicating the score for the assistant.
In the subsequent line, please provide a comprehensive explanation of your evaluation, avoiding any potential bias.
\end{tcolorbox}

\end{document}